\documentclass[10pt,twocolumn,letterpaper]{article}

\usepackage{cvpr}              %

\usepackage[dvipsnames]{xcolor}
\usepackage{times}
\usepackage{amsmath}
\usepackage{amssymb}
\usepackage{booktabs}
\usepackage{caption}
\usepackage{cuted}
\usepackage{graphicx}
\usepackage[numbers,sort]{natbib}

\usepackage[accsupp]{axessibility} %

\def\eg{\emph{e.g.}} 
\def\ie{\emph{i.e.}}

\DeclareMathOperator*{\argmax}{arg\,max}

 \newcommand{\supp}{the Appendix\xspace}

\newif\ifdark
\IfFileExists{/Users/vedaldi/.bash_profile}{
\immediate\write18{%
if defaults read -g AppleInterfaceStyle 2>/dev/null;
then echo \\darktrue > /tmp/displaymode.tex;
else echo \\darkfalse > /tmp/displaymode.tex; fi}
\input{/tmp/displaymode.tex}
\ifdark
\definecolor{pcolor}{HTML}{1E1E1E}
\definecolor{tcolor}{HTML}{C5C5C5}
\else
\definecolor{pcolor}{HTML}{FDF6E3}
\definecolor{tcolor}{HTML}{333333}
\fi
\pagecolor{pcolor}
\color{tcolor}
\hbadness=\maxdimen
\vbadness=\maxdimen
\vfuzz=30pt
\hfuzz=30pt
}{} 
\usepackage{pifont}%
\newcommand{\cmark}{\ding{51}}%
\newcommand{\xmark}{\ding{55}}%

\usepackage{colortbl}
\usepackage{amsmath}

\definecolor{Gray}{gray}{0.85}
\definecolor{LightCyan}{rgb}{0.88,1,1}

\newcolumntype{a}{>{\columncolor{Gray}}c}

\newcommand{\midsepchange}{\aboverulesep = 0.15mm \belowrulesep = 0.4mm}
\midsepchange
\usepackage[pagebackref,breaklinks,colorlinks]{hyperref}
\usepackage{cleveref}

\def\cvprPaperID{11391} %
\def\confName{CVPR}
\def\confYear{2022}

\makeatletter
\renewcommand{\paragraph}{%
  \@startsection{paragraph}{4}%
  {\z@}{.5ex \@plus 1ex \@minus .2ex}{-1em}%
  {\normalfont\normalsize\bfseries}%
}
\makeatother

\title{Self-supervised object detection from audio-visual correspondence}
\author{
  Triantafyllos Afouras$^{1}$\thanks{Joint first authors.}\,\,\thanks{Work done during an internship at FAIR.} \quad
  Yuki M. Asano$^{2}$\footnotemark[1] 
  \quad {Francois Fagan}$^3$ 
  \quad {Andrea Vedaldi}$^3$ 
  \quad {Florian Metze}$^{3}$  \vspace{0.6em}
\\
   $^1$ University of Oxford \hfill $^2$ University of Amsterdam \hfill $^3$ Meta AI \vspace{0.6em}
   \\
      \texttt{afourast@robots.ox.ac.uk}}

\begin{document}
\maketitle
\setlength{\belowcaptionskip}{-0.75em}

\begin{abstract}
We tackle the problem of learning object detectors without supervision.
Differently from weakly-supervised object detection, we do not assume image-level class labels.
Instead, we extract a supervisory signal from audio-visual data, using the audio component to ``teach'' the object detector.
While this problem is related to sound source localisation, it is considerably harder because the detector must classify the objects by type, enumerate each instance of the object, and do so even when the object is silent.
We tackle this problem by first designing a self-supervised framework with a contrastive objective that jointly learns to classify and localise objects.
Then, without using any supervision, we simply use these self-supervised labels and boxes to train an image-based object detector.
With this, we outperform previous unsupervised and weakly-supervised detectors for the task of object detection and sound source localization.
We also show that we can align this detector to ground-truth classes with as little as one label per pseudo-class, and show how our method can learn to detect generic objects that go beyond instruments, such as airplanes and cats.
\end{abstract}

\section{Introduction}

While recent progress in learning image and video representations has been substantial~\cite{wu18unsupervised,he2019moco,chen2020simple,grill20bootstrap}, this has not yet translated into an ability to learn interpretable and actionable concepts automatically.
By that, we mean that some manual labels are still required in order to map unsupervised representations to useful concepts such as image classes or object detections.
In this paper, we thus consider the problem of learning interpretable concepts without any manual supervision.
In particular, we focus on a problem that has not been explored extensively in the literature: learning to simultaneously detect and classify objects with no manual labels at all.

\begin{figure}[!t]
  \includegraphics[width=\columnwidth]{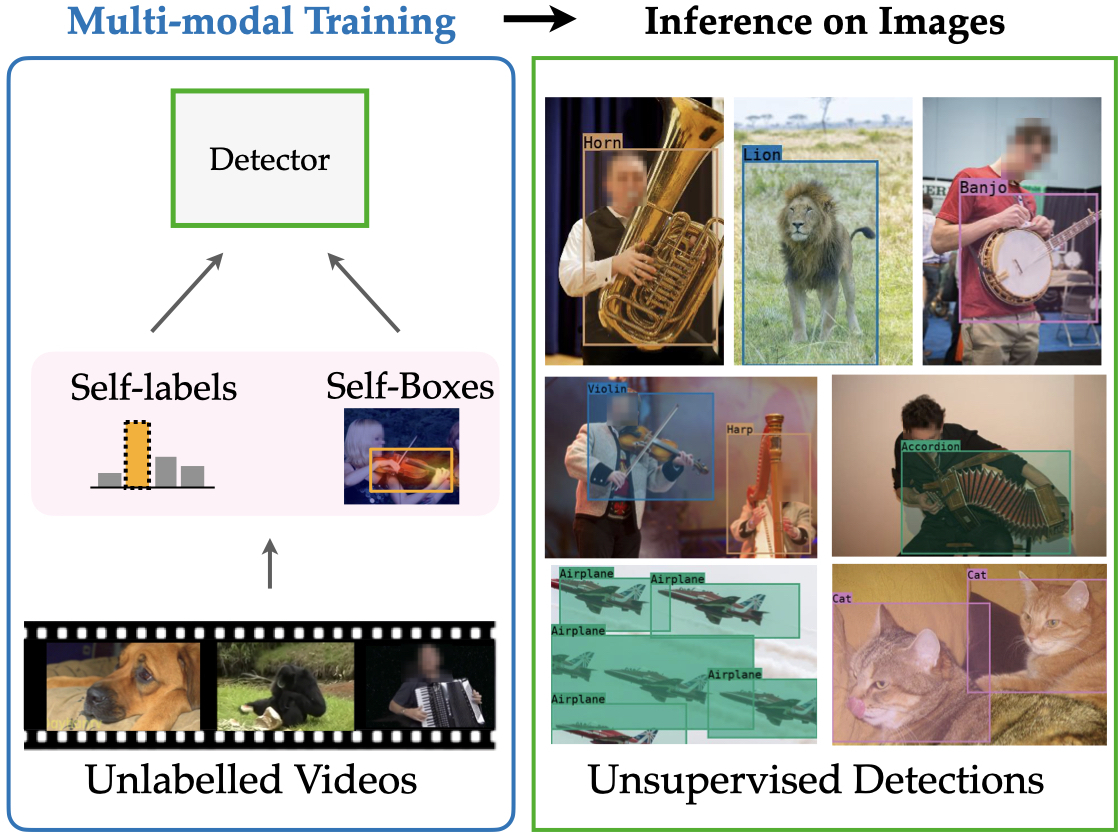}
  \caption{ \textbf{We train an object detector simply by watching videos.} Without using any manual annotations, we learn to detect different objects in images, by first self-labelling boxes and object categories and then using those as targets to teach a detector.  
  The detection results shown are outputs from our trained model; for visualisation purposes we show Hungarian-matched labels. 
  \label{fig:splash}}
\end{figure}

This problem is related to weakly supervised object detection (WSOD~\cite{nguyen09weakly,bilen11object}), with the difference that, in WSOD, the learning algorithm is given image-level labels telling it whether the image contains an occurrence of a given object type or not.
Inspired by recent work in self-supervised learning, we seek to replace this source of external supervision with an internal supervisory signal extracted from the observation of video data.
Videos are far richer than images, for example because they contain motion.
Here, we focus on the multi-modal aspect of videos and use sound as a weak and noisy cue to learn about objects in the visual component of the data.

The power of multi-modal self-supervision has been demonstrated before in self-supervised representation learning, and, closely related, in \emph{video clustering}~\cite{asano20labelling}.
However, while video clustering can provide an interpretation of the data in terms of discrete classes, it does not provide any information about the location of the relevant objects in images.
On the other hand, \emph{sound source localisation}~\cite{kidron05pixels,owens16visually,arandjelovic18objects,aytar16soundnet:} has considered precisely the problem of localizing the source of sounds in images.
It is therefore tempting to trivially combine image classification and sound source localisation in the hope of learning the type and location of objects automatically.

Unfortunately, such an approach does \emph{not} lead to a satisfactory object detector.
To understand why, it is important to note that the goal of sound source localisation is to \emph{localize the sound while it is being heard}.
This is insufficient for a detector because many objects emit sounds only occasionally and they become invisible to source localisation when they are silent.
Instead, a detector that works in the visual domain should be responsive even when the object cannot be heard.
Furthermore, source localisation methods generally only extract a heatmap giving the distribution of possible object locations; in contrast, a detector solves the much harder problem of enumerating all individual object instance that occur in an image by outputting instance-specific bounding boxes.

In order to solve these issues, we should treat the sound component as a useful cue to \emph{learn} an object detector, but not as a cue which is \emph{necessary for detection}.
Instead, we consider the problem of taking as input a collection of raw videos and producing a list of object classes and locations, in order to train an image-based detector.

On a high level, our method is based on the following observation:
we can use a sound source localisation network to learn about possible locations of sounding objects in videos.
From this, we can extract a collection of bounding box pseudo-annotations for the objects and use those to learn a standard object detector.
Because the latter only uses the visual modality, it immediately transfers to the detection of objects even when no relevant sound is present.

However, one challenge is that sound source localisation does not provide the necessary class information to train class-specific detectors, effectively resulting in only learning a region proposal network for generic objects, with high rates of false positives.
To this end, we note that most sound source localizers are based on noise-contrastive formulations that, together with clustering-based approaches, are currently state-of-the-art in self-supervised representation learning.
From this, we derive a joint formulation that can simultaneously benefit from and learn to localize sound sources and classify them without \textit{any} supervision.
The resulting output can then be used to train any off-the-shelf object detector such as a Faster-RCNN~\cite{ren16faster} to learn an object detector without any supervision, as shown in Fig.~\ref{fig:splash}.

Empirically, we test our method by training and testing on VGGSound~\cite{chen20vggsound} and  AudioSet~\cite{gemmeke17audio}, as well as testing only on a subset of OpenImages~\cite{kuznetsova2018open}.

\begin{figure*}[ht]
\centering
\includegraphics[width=0.93\textwidth]{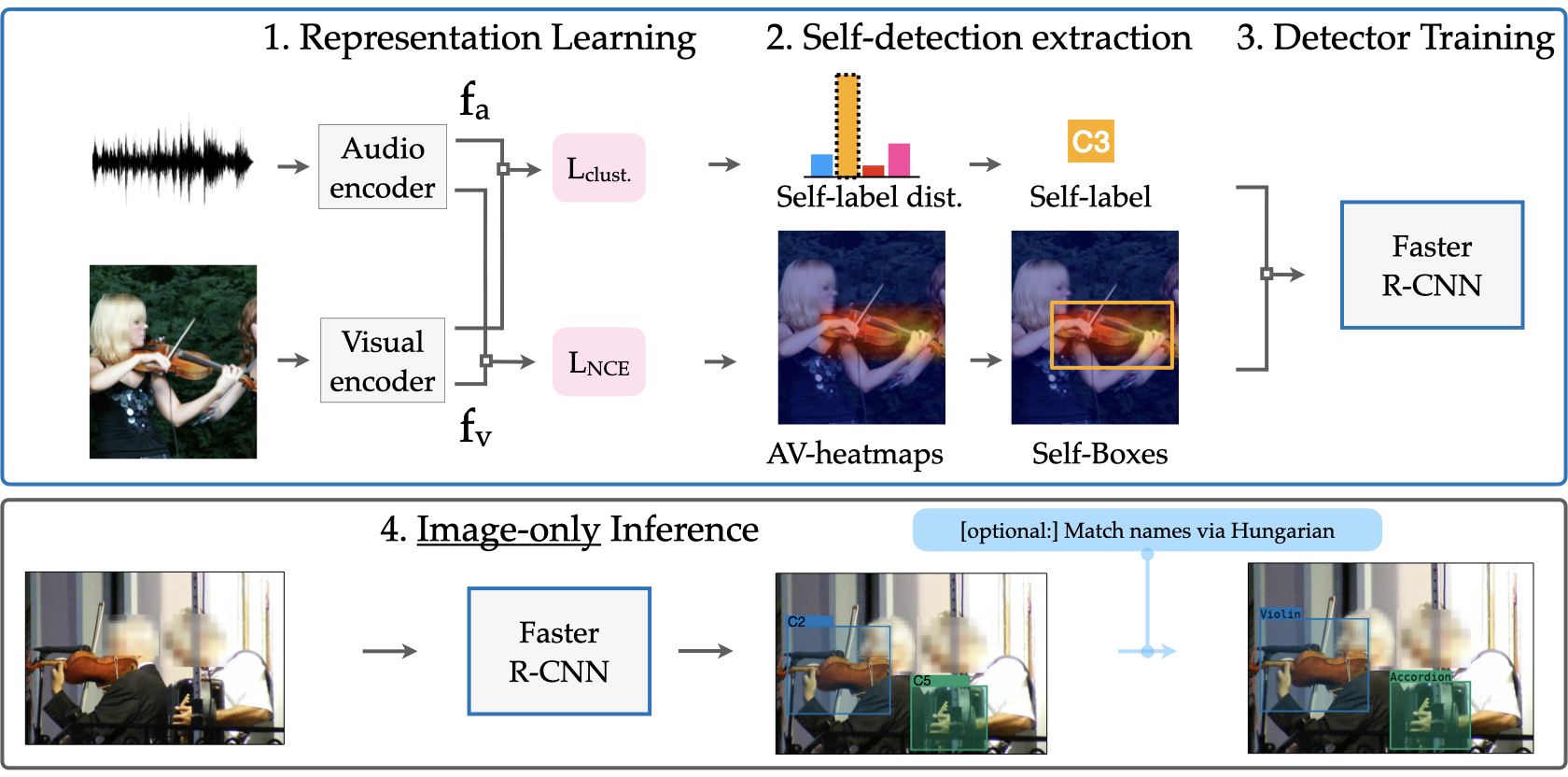}
\caption{Self-supervised object detection from audio-visual correspondence: We combine noise-constrastive and clustering-based self-supervised learning to generate self-detections (boxes and labels) and use those as targets to train a detector. The trained detector can be used to detect objects from many categories on images without requiring audio.
}\label{fig:pipeline}
\end{figure*}
\section{Related Work}

\paragraph{Audio-Visual Sound Source Localisation.}

Early work in sound-source localisation includes probabilistic models for localisation~\cite{hershey1999audio,fisher2000learning,kidron05pixels} and segmentation~\cite{izadinia2012multimodal}, but more recently the focus has shifted to dual-stream neural networks.
For example,~\cite{arandjelovic18objects,senocak2018learning,harwath2018jointly} propose a contrastive learning approach that matches the visual and audio components of the data.
The work of~\cite{Hu_2019_CVPR,Hu2020CurriculumAL} instead clusters visual and audio features, associating to them centroids by means of a contrastive loss.
Other works~\cite{Owens2018b,Afouras20b} learn heatmaps by exploiting audio-visual synchronization
in the same video, used previously for lip-to-mouth synchronization and
active-speaker detection~\cite{Marcheret15,chung16},  or by leveraging explicit attention modules~\cite{khosravan2018attention}.
Zhao et al.~\cite{zhao2019sound,zhao2018sound} learn to associate pixels with audio sources by training with a mix-and-separate objective.
Others~\cite{qian2020multiple} combine activation maps learned from class labels~\cite{selvaraju2017grad,chattopadhay2018grad} with a contrastive objective, use different levels of supervision and fusion techniques~\cite{Ramaswamy_2020_WACV},
or improve heatmaps by mining hard negative locations~\cite{chen2021localizing}.

The work most similar to ours is~\cite{hu20discriminative}, who first train a source localisation model with a contrastive objective and then use the learned heatmaps to extract object
representations that are clustered using K-means. The cluster assignments are then used to train
classifiers on top of the audio and video encoders.
The paper proceeds to use these learned representations to discriminatively localize sound sources
while suppressing quiet objects in `cocktail party' scenarios.

Compared to our work, none of the above can detect and thus enumerate individual object occurrences because they produce heatmaps.
Furthermore, they all require audio during inference, and therefore cannot be used on individual images or to detect silent objects.

\paragraph{Audio-visual category discovery.}

Learning visual categories is usually cast as image clustering, for which there is abundant prior work, such as recent `deep clustering' methods~\cite{ji2019invariant, caron2018deep,asano20self-labelling,gansbeke2020learning,yang2016joint, xie2016unsupervised}, or clustering with segmentation~\cite{vangansbeke2020unsupervised}.
However, there is less work for clustering audio-visual data.
In~\cite{alwassel_2020_xdc} the authors extend Deep Cluster~\cite{caron2018deep} to the video domain by constructing two sets of labels from opposing modalities, which are used for cross-modal representation learning.
The work of~\cite{rouditchenko2019self} combines clustering with audio-visual co-segmentation achieving combined audio-visual source separation.
In~\cite{asano20labelling}, the authors extend the self-labelling method of~\cite{asano20self-labelling} to multi-modal data by learning a shared set of labels between the two modalities.
This work builds on the latter to complement and boost sound source localisation in a joint learning framework.

\paragraph{Weakly Supervised Object Detection (WSOD).}

Weakly supervised detection uses (manual) image-level category labels without bounding box annotations. 
Many approaches are based on a form of multiple-instance learning~\cite{bilen16weakly,tang17multiple,wan2019cmil,zeng2019wsod2,yan19cmidn,Zhang18w2f,yang2019activity,Gonthier_2019,xie2021online,meng2021foreground}, or proposal clustering \cite{tang18pcl:}. 
Recent works in the area~\cite{jie2017deep,ren20instance-aware} combine a variety of ideas, such as self-training~\cite{zou19confidence} and spatial dropout~\cite{wang17a-fast-rcnn:} or explore the use of mixed annotations~\cite{ren20ufo2:}.
Other works obtain improvements by adding curriculum learning \cite{zhang2018zigzag}, using motion cues in videos~\cite{singh-cvpr2019}, adversarial training~\cite{Shen18ganWSOD},
combining segmentation and detection~\cite{ge2018multievidence,shen19cyclic,li2019weakly}, or
modelling the uncertainty of object locations \cite{arun2018dissimilarity}.
Other methods use technique such as CAM or analogous techniques~\cite{zhou2016,bazzani2016self,selvaraju2017grad,chattopadhay2018grad,singh2017hide,fong2019explanations} as a form of weakly supervised saliency or localisation maps.
Recent works have suggested that saliency methods can also be applied to self-supervised networks~\cite{gur2020visualization,baek20psynet:}, \eg for object co-localisation~\cite{baek20psynet:}.

\paragraph{Self-supervised multi-modal learning.}

Our work is also related to methods that use multiple modalities for representation learning~\cite{owens16ambient,aytar16soundnet,arandjelovic17look,morgado20learning,alwassel_2020_xdc,asano20self-labelling,patrick2020multi} and synchronization~\cite{chung16out-of-time:,korbar18cooperativeLearning,owens18audio-visual}.
A number of recent papers have leveraged speech as a weak supervisory signal to train video representations~\cite{sun19videobert:,sun19contrastive,miech19end-to-end,li20learning,nagrani20speech2action:}
whereas~\cite{alayrac20self-supervised} uses speech, audio and video.
Some works distil knowledge learned from one modality into another~\cite{Zhao18radiopose,gan2019selfsupervised,Albanie18,Afouras20}.
Other works incorporate optical flow and other modalities~\cite{zhao2019sound,piergiovanni20evolving,han20self-supervised,han20memory-augmented} to learn representations.
For instance, the work of~\cite{tian18audio-visual} learns to temporally localize audio events through audio-visual attention.
CMC~\cite{tian2019contrastive} learns representations that are invariant to multiple views of the data such as different color channels.
Multi-modal self-supervision is also used to learn sound source separation in~\cite{gao19co-separating}, albeit they assume to have pre-trained detectors.

\section{Method}\label{s:method}

Our goal is to learn object detectors using only unlabeled videos, simultaneously learning to enumerate, localize and classify objects.
Our approach consists of three stages summarized in Fig.~\ref{fig:pipeline}:
first, we learn useful representations using clustering and contrastive learning;
second, we extract bounding boxes and class categories by combining the trained localisation and classification networks;
third, we train an off-the-shelf object detector by using these self-extracted labels and boxes as targets.
Next, we explain each stage and refer the reader to the \supp for further architecture and training details.

\subsection{Representation Learning}

\paragraph{Sound source spatial localisation.}

Our method starts by training a sound source localisation network (SSLN) using a contrastive learning formulation inspired by~\cite{arandjelovic18objects}.
The SSLN is learned from pairs $(v,a)$, where
$
v \in \mathbb{R}^{3 \times H \times W}
$
is a  video frame (\ie, a still image) and 
$
a \in \mathbb{R}^{T \times F}
$
is the spectrogram of the audio captured in a temporal window centered at that particular video frame.

We consider a pair of deep neural networks.
The first network
$
f^v(v) \in \mathbb{R}^{C\times h \times w}
$
extracts from the video frame a field of $C$-dimensional feature vectors, one per spatial location.
We use the symbol $f^v_u(v) \in \mathbb{R}^C$ to denote the feature vector associated to location $u\in\psi = \{1,\dots,h\} \times \{1,\dots,w\}$.
Here $h\times w$ is the resolution at which the spatial features are computed and is generally a fraction of the video frame resolution $H \times W$.
The second network
$
f^a(a) \in \mathbb{R}^{C}
$
extracts instead a feature vector for the audio signal.

Importantly, the spatial and audio features share the same $C$-dimensional embedding space and can thus be contrasted.
We further assume that the vectors $f^v_u(v)$ and $f^a(a)$ are $L^2$ normalized (this is obtained by adding a normalization layer at the end of the corresponding networks).
The cosine similarity of the two feature vectors is then used to compute a heatmap of spatial locations, with the expectation that objects that are correlated with the sounds would respond more strongly.
This heatmap is given by:
$$
 h_u(v,a)
 =
 \langle f^v_u(v),f^v(a) \rangle/\rho,
 ~~~
 u \in \psi,
$$
where $\rho$, is a learnable temperature parameter.

\newcommand{\B}{\mathcal{\mathcal{B}}}

For the multi-modal contrastive learning formulation~\cite{chung2019perfect,patrick2020multi,oord2018representation}, the heatmap is converted in an overall score that the video $v$ and audio $a$ are in correspondence.
This is done by taking the maximum of the response:
$$
S(v,a) =  \max_{u\in\psi} h_u(v,a).
$$
The contrastive learning objective is defined by considering videos $(v,a)\in\B$ in a batch $\B$.
This comprises two terms.
The first tests how well a video frame matches with its specific audio among the ones available in the batch:
$$
\mathcal{L}_{a\rightarrow v}(\B)
=
-
\frac{1}{|\B|}
\sum_{(v,a)\in\B}
\log \frac
{
\exp S(v,a)
}
{
\sum_{(v',a')\in\B}
\exp S(v,a')
}.
$$
The second is analogous, testing how well an audio matches with its specific video frame:
$$
\mathcal{L}_{v\rightarrow a}(\B)
=
-
\frac{1}{|\B|}
\sum_{(v,a)\in\B}
\log \frac
{
\exp S(v,a)
}
{
\sum_{(v',a')\in\B}
\exp S(v',a)
}.
$$
These two losses are averaged in the \emph{noise-contrastive} loss:
\begin{equation}\label{e:loss-nce}
\mathcal{L}_\text{NC}(\B) =
(
  \mathcal{L}_{a \rightarrow v}(\B) +
  \mathcal{L}_{v \rightarrow a}(\B)
)/2
\end{equation}

\paragraph{Category self-labeling.}

Spatial localisation does not provide any class information, whereas our goal is to also associate `names' to the different objects in the dataset.
To this end, we consider the self-labelling approach of~\cite{asano20labelling}.
To briefly explain the formulation, let $y(v,a) \in \mathcal{Y}=\{1,\dots,K\}$ be a label 
associated to the training pair $(v,a)$.
We also consider two classification networks.
The first maps a video $v$ to class scores $g^v(v)\in\mathbb{R}^K$ and is optimized by minimizing the standard cross-entropy loss:
$$
\mathcal{L}_v(\B|y)
=
-
\frac{1}{|\B|}
\sum_{(v,a)\in\B}
\log \operatorname{softmax}(y(v,a) \, | \, g^v(v)).
$$
Note that this classification loss is equivalent to a contrastive loss on the cluster indices (as opposed to image indices) without normalization:  
As the last classification layer can be viewed as computing dot-products with each corresponding cluster's feature,  it pushes the representation towards the feature of the corresponding cluster and away from the other clusters.
The other network $g^a(a)$ is analogous, but uses the audio signal:
$$
\mathcal{L}_a(\B|y)
=
-
\frac{1}{|\B|}
\sum_{(v,a)\in\B}
\log \operatorname{softmax}(y(v,a)\, | \,g^a(a)).
$$
As noted in~\cite{asano20labelling}, the crucial link between the two losses is that the labels $y$ are shared between modalities.
This is obtained by averaging the two losses:
\begin{equation}\label{e:loss-ce}
  \mathcal{L}_\text{clust}(\B|y) =
  (
    \mathcal{L}_{v}(\B|y) +
    \mathcal{L}_{a}(\B|y)
  )/2.
\end{equation}
Note that the labels $y$ are unknown; following~\cite{asano20labelling} these are learned in an alternate fashion with the classification networks, minimizing the same loss~\eqref{e:loss-ce}.
In order to avoid degenerate solutions, the labels' marginal distribution must be specified, e.g. using a simple equipartitioning constraint:
\begin{equation}\label{e:equi}
\frac{1}{|\mathcal{D}|}
\sum_{(v,a)\in\mathcal{D}}
1_{\{y(v,a) = k\}}
=
\frac{1}{K}
~~~\text{for all~}
k=1,\dots,K
\end{equation}
where $\mathcal{D}$ denotes the entire dataset (union of all batches).
Optimizing $y$ can be done efficiently by using the SK algorithm as in~\cite{asano20labelling}. 

\paragraph{Joint training.}

To summarize, given the dataset $\mathcal{D}$, we optimize stochastically w.r.t.~random batches $\mathcal{B}$ the loss:
\begin{equation}\label{e:loss-joint}
  \mathcal{L}(\mathcal{B}|y)
  = 
  \lambda \mathcal{L}_\text{NC}(\mathcal{B})
  +
  (1-\lambda) \mathcal{L}_\text{clust}(\mathcal{B}|y)
\end{equation}
where $\lambda$ is a balancing hyperparameter.

The loss is optimized with respect to the localisation networks $f^v$ and $f^a$ and the classification networks $g^v$ and $g^a$.
These networks share common backbones $q^v$ and $q^a$ and differ only in their heads, so they can be written as
$f^v = \hat f^v \circ q^v$,
$g^v = \hat g^v \circ q^v$,
$f^a = \hat f^a \circ q^a$ and
$g^a = \hat g^a \circ q^a$.

The model is trained by alternating between updating the labels $y$ with \cref{e:loss-ce} under constraint~\eqref{e:equi} and updating the networks by optimizing \cref{e:loss-joint}.

\subsection{Extraction of Self-labels for Detection}

Once the localisation and classification networks have been trained, they can be used to extract self-annotations for training a detector.
This is done in two steps:
extracting object bounding boxes and finding their class labels.

\paragraph{Box extraction.}

To obtain ``self-bounding boxes'' for the objects, we use the simple heuristic suggested by~\cite{zhou2016}:
the heatmap $h(v,a)$ is thresholded at a value $\epsilon(h)$, the largest connected component is identified, and a tight bounding box $t^*(v,a)\in\Omega^2$ around that component is extracted (the notation means that the box is specified by the location of the top-left and bottom-right corners).

The threshold is determined dynamically as a convex combination of the maximum and average responses of the heatmap, controlled by hyperparameter $\beta$:
\begin{equation}\label{e:heatmap_threshold}
\epsilon(h) 
=
\beta \max_{u\in\psi} h_{u} 
+ (1-\beta) \frac{1}{|\psi|} \sum_{u\in\psi} h_{u}.
\end{equation}

\paragraph{Class labelling.}

As noted above, we only extract a single object from each frame for the purpose of training the detector.
Likewise, we only need to extract a single class label for the frame.
This is done by taking the maximum response of the visual and audio-based classification networks:
\begin{equation}\label{e:class_label}
y^*(v,a) = \argmax_{y\in\mathcal{Y}}
 [g^v_y(v) + g^a_y(a)].
\end{equation}

\paragraph{Filtering the annotations.}
The assumption that frames contain a dominant object introduces noise but simplifies the problem and gives us the ability to use the audio to obtain purer clusters.
Notably, we do not require the method above to work for \emph{all} frames but instead rely on our detector to smooth over the specific and noisy self-annotations to learn a holistic detection.

\subsection{Training the Object Detector}

The process described above results in a shortlist of training triplets $(v,t^*,y^*)\in\mathcal{D}_\text{det}$, where $v$ is a video frame (an image), $t^*$ is the extracted bounding box and $y^*$ is its class label.
We use this dataset to train an off-the-shelf detector, in particular Faster R-CNN~\cite{ren15faster} for its good compromise between speed and quality.

Recall that, given an image $v$, Faster R-CNN detector considers a set of bounding box proposals $m \in M(v) \subset \Omega^2$.
It then trains networks
$
y(m) = f_\text{det}^\text{cls}(m|v)
\in \{1,\dots,K,\text{bkg}\}
$ 
and
$
t(m) = f_\text{det}^\text{loc}(m|v)
\in \mathbb{R}^4
$
inferring, respectively, the class label $y(m)$ and a refined full-resolution bounding box $t(m)$ for the box proposal $m$.
The label space is extended to also include a \emph{background class} $\text{bkg}$, which is required as most proposals do not land on any object.

The detector is trained by finding an association between proposals and annotations.
To this end, if $m^* = \argmax_{m\in M(v)} \operatorname{IoU}(m,t^*)$ is the proposal that matches the pseudo-ground truth bounding box $t^*$ the best, one optimizes:
\begin{multline*}
\mathcal{L}_\text{det}(v,t^*,y^*)
=
\mathcal{L}_\text{reg}(t(m^*),t^*)
+
\mathcal{L}_\text{cls}(y(m^*),y^*) \\
+
\sum_{m \in M(v): \operatorname{IoU}(m,t^*) < \tau} 
\mathcal{L}_\text{cls}(y(m),\text{bkg}).
\end{multline*}
Here $\mathcal L_\text{reg}$ is the $L^1$ loss for the bounding box corner coordinates and $\mathcal{L}_\text{cls}$ the standard cross-entropy loss.
Intuitively, this loss requires the best proposal $m^*$ to match the pseudo-ground truth class $y^*$ and bounding box $t^*$ of, while mapping proposal $m$ that are a bad match ($\tau\leq 0.7$) to class $\text{bkg}$.
Further details, including how the region-proposal network that generates the proposals is trained, are given in the \supp

\paragraph{Discussion.}

Training a detector is obviously necessary to solve the problem we set out to address.
However, it can also be seen as a way of extracting `clean' information from the noisy self-annotations.
Specifically:
(i) the noise in individual annotations is smoothed over the entire dataset;
(ii) because of the built-in NMS step, the detector still learns to extract multiple objects per image even though a single self-annotation is given for each training image;
(iii) by learning to reject a large number of false bounding box proposals, the detector learns to be more precise than the self-annotations.

\section{Experiments}\label{s:experiments}

We first introduce the datasets, experimental setup and relevant baselines; we then test our method against those, analyse it further via ablations and its capacity to generalize.

\subsection{Datasets}\label{s:datasets}

\paragraph{AudioSet-Instrument.} AudioSet~\cite{gemmeke17audio} is a large scale audio-visual dataset consisting of 10-second video clips originally from YouTube.
For training we use the \textit{AudioSet-Instruments}~\cite{arandjelovic18objects} subset of the ``unbalanced'' split, containing 110 sound source classes as well as its more constrained subset used by~\cite{hu2020discriminative} spanning 13 instrument classes.
Following previous work we use the ``balanced'' subset for evaluation on the annotations provided by~\cite{hu2020discriminative}.

\paragraph{VGGSound.}
VGGSound contains over 200K 10-second clips from YouTube spanning 309 categories of objects where there is some degree of correlation between the audio and the video. 
We create one subset by keeping only the 50 musical instrument categories yielding around 54K training videos, and one other subset, by keeping from those only the 39 categories that can be roughly mapped to the test-set annotations (details in \supp). 
For VGGSound pseudo-ground truth test-set annotations are obtained using a supervised detector from~\cite{gao19co-separating}, following~\cite{hu2020discriminative}.

\paragraph{OpenImages.} For evaluation, we also use the subset of the OpenImages~\cite{kuznetsova2018open} dataset containing musical instruments, which spans 15 classes. 
\begin{figure*}[!htb]
    \includegraphics[width=\textwidth]{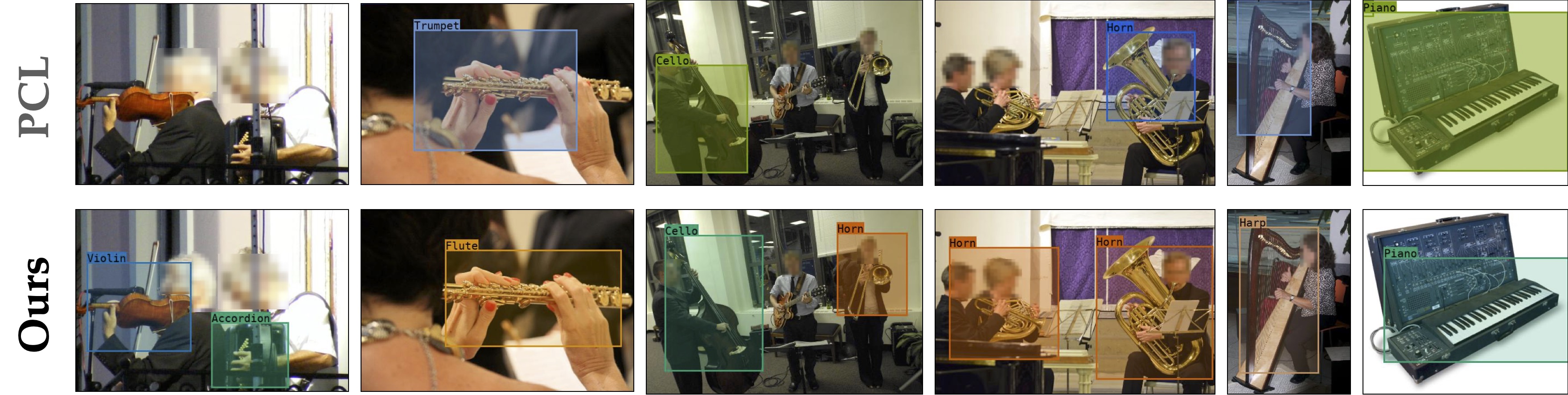}
    \caption{\textbf{Qualitative results and comparison} with a weakly supervised object detection method, PCL~\cite{tang18pcl:}, on the OpenImages test set.
    Our method accurately detects objects, capturing their boundaries, even though it has been trained without \textit{any} supervision. 
    For visualisation purposes, we show the labels obtained from matching with the Hungarian method. 
    More qualitative results provided in \supp.
     \label{fig:qualitative}}
\end{figure*}

\subsection{Baselines}

There is currently no prior work on learning an object detector for multiple object classes without any supervision.
Instead, we compare against weakly-supervised detectors (hence using image-level labels) and unsupervised localisation methods that only produce heatmaps (not detections).

For \textbf{weakly-supervised detection}, we consider PCL~\cite{tang18pcl:}, the strongest such baseline for which we could find an implementation.

For our second baseline, we consider \textbf{heatmap-based localisation methods} that, similarly to us, use cross-modal self-supervised learning.
Here we compare against the state-of-the art DSOL method of~\cite{hu2020discriminative} that produces a heatmap roughly localizing the objects and producing class pseudolabels.

Finally, we also compare against \textbf{other baselines} such as simply predicting a large centered box and class-agnostic region proposal methods such as selective search, and using a RPN obtained from supervised training on COCO~\cite{lin14microsoft}.
Further details are provided in the \supp.

\begin{figure*}
\centering
\begin{minipage}[b]{.65\textwidth}
  \centering
        \footnotesize
            \setlength{\tabcolsep}{1pt}
            \centering
                \setlength{\tabcolsep}{1pt}
                \begin{tabular}{l  l  @{\hskip 0.1in} ccc @{\hskip 0.1in} ccc@{\hskip 0.1in} ccc}
                    \toprule
                                        &         &  \multicolumn{3}{c}{\textbf{VGGSound}} & \multicolumn{3}{c}{\textbf{Audioset}} & \multicolumn{3}{c}{\textbf{OpenImages}} \\
             Method \qquad  No labels?$_\searrow$             & {}         & mAP$_{30}$ & mAP$_{50}$ & mAP    & mAP$_{30}$ & mAP$_{50}$ & mAP & mAP$_{30}$ & mAP$_{50}$ & mAP  \\
            \midrule
            PCL (WSOD)~\cite{tang18pcl:}       & {\textcolor{red}\xmark}     &  {54.9}  & 27.7 & 7.6     &  39.0 & 17.5  &   4.4 &       {37.9} & 14.5 &  3.5  \\ 
            Ours - weak sup.                   & {\textcolor{red}\xmark}      & 67.6 &  42.9 &  14.2  & 50.6 &  30.9 &  10.3 & 48.9 &  33.7 &  9.5 \\
            \midrule
            \midrule
            Center Box*                         & {\textcolor{green}\cmark}     &  29.6  & 5.6 & 1.5    &  15.1 & 3.5  &  0.7 &        20.7 & 4.2 &  0.8  \\
            Selective Search* ~\cite{Uijlings13}          & {\textcolor{green}\cmark}     &  5.2  & 1.1 & 0.4     &  2.8 & 0.4  &  0.1 &         7.4 & 2.1 &  0.7   \\
            COCO-trained RPN*                  & {\textcolor{red}\xmark}     &  33.4  & 7.5 & 1.6    &  19.0 & 4.1  &  0.8 &                24.4 & 11.1 &  2.6   \\
            \midrule
            Ours - self-boxes*                & {\textcolor{green}\cmark}      & 48.1  & 29.6 & 10.0    &  27.8 & 14.1  & 4.8  &      NA & NA &  NA                                 \\
            \textbf{Ours - full }                          & {\textcolor{green}\cmark} & \textbf{52.3} &  \textbf{39.4} &  \textbf{14.7}  & \textbf{44.3} &  \textbf{28.0} &  \textbf{9.6}  & \textbf{39.9} &  \textbf{28.5} &  \textbf{7.6} \\
            \bottomrule
            \end{tabular}
            \vspace{-0.5em}
            \captionof{table}{
            \textbf{Self-supervised object detection.} We report object detection metrics across three test datasets and find our method is far superior to other unsupervised approaches and outperforms even the weakly supervised baseline in most metrics. 
            For methods denoted by *, we report class-agnostic evaluation numbers. 
            The class-agnostic performance of the self-boxes that are used to train the detector reveals that the latter greatly outperforms them, which highlights the benefit of our approach. \label{tab:results}
            }
\end{minipage}%
\hfill%
\begin{minipage}[b]{.32\textwidth}
    \footnotesize
        \centering
            \setlength{\tabcolsep}{1pt}
            \begin{tabular}{l  ccc}
            \toprule
                                                          & \multicolumn{2}{c}{\textbf{single-instr.}} & \textbf{multi-instr.} \\
                 Method                                   &  IoU-0.5   & AUC &  cIoU-0.3 \\
                \midrule
                Sound of pixels~\cite{zhao2018sound} & 38.2 & 40.6 & 39.8\\
                Object t. Sound~\cite{arandjelovic18objects} & 32.7 & 39.5  & 27.1\\
                Attention~\cite{senocak2018learning} & 36.5 & 39.5 & 29.9\\
                DMC~\cite{hu19deep}  & 32.8 & 38.2 & 32.0\\
                DSOL~\cite{hu2020discriminative}    &  38.9 & 40.9 & 48.7 \\
                \midrule
                \textbf{Ours}                  &  \textbf{50.6} & \textbf{47.5} & \textbf{52.4} \\ 
                \bottomrule 
            \end{tabular}
        \vspace{-3pt}
        \captionof{table}{\textbf{Comparison to sound localisation methods.} 
        Since our detector does not require audio, we obtain detections on the video frames directly. Our model outperforms the  baselines.
        Baselines numbers taken from~\cite{hu20discriminative}. 
        \label{tab:ciou}}

\end{minipage}
\end{figure*}

\subsection{Implementation Details.}

\paragraph{Assessing class pseudolabels.}
Since class pseudolabels do not come with the name of the class (they are just cluster indices), they must be put in correspondence with human-labelled classes for evaluation.
Following prior work in unsupervised image clustering~\cite{bautista2016cliquecnn, gansbeke2020learning,ji2019invariant,asano20self-labelling}, we apply Hungarian matching~\cite{kuhn1955hungarian} to the learned clusters and the ground truth classes.
Importantly, the matching is done \textit{after} the detector is trained and only done for assessment; meaning that the detector does not use any manual label.

\paragraph{Detector training.}
If not stated otherwise, the localizer and detector are trained on VGGSound and AudioSet whereas OpenImages are only used for evaluation.
We do not have any information on the number of instruments in VGGSound and use all videos with no single/multi-object curation.
For a fair comparison with DSOL, and only for the relevant experiment in \Cref{tab:ciou}, we train on AudioSet using the single-instrument subset for learning the localizer.

\paragraph{Number of clusters.}
For VGGSound training we use $K=39$ if not stated otherwise, matching the 39 object types in the training set. 
Since the dataset is roughly balanced, uniform marginals are used as described in~\cite{asano20labelling}.
For AudioSet training we use $K=30$ and Gaussian marginals. 
Further implementation details can be found in the \supp

\subsection{Results}

\paragraph{Self-supervised object detection.}

We summarise the results of our evaluations on the three test sets that we consider in \Cref{tab:results}.
Following the image object detection literature, we use mAP at different IoU thresholds as the evaluation metric. 
Our method clearly outperforms the PCL baseline even though it uses no manual annotations at all during training. 
PCL outperforms our approach in some of the datasets only if the IoU  threshold used for mAP computation is relaxed substantially (0.3 IoU).
However, for stricter thresholds our approach works better, which suggests that our detections have a relatively high spatial accuracy.

To understand the impact of the noisy class self-labels, we also train and test a detector (Ours - weak sup.) with the bounding box labels from our localisation network, but utilising the ground truth video categories. 
The resulting performance difference is modest, resulting in a 3\% AP50 drop in VGGSound and AudioSet.
This further demonstrates the accuracy of our class self-labels, but also shows the potential of our model to leverage weak supervision if available.  

\paragraph{Per-class performance breakdown.}
\begin{table*}[tb]
\centering
\setlength{\tabcolsep}{2.5pt}
\footnotesize
        \begin{tabular}{l c cccccccccccccc}
        \toprule
        Dataset & \textbf{mAP$_{50}$} & Accordion  & Cello & Drum & Flute & Horn & Guitar & Harp   & Piano & Saxophone & Violin & Banjo & Trombone & Trumpet  & Oboe  \\
        \midrule
        OpenImages  & 28.5 &  75.3   & 30.2  &  6.6  &   6.5  &  15.0 & 14.5 &  80.4 & 28.8  &  22.5   & 28.8 &   57.0    &   9.7   &  18.1 &  6.3  \\
        Audioset    & 28.0 &  41.3   &  44.9 &  0.9   &    5.5  &  21.7 & 39.5 &  82.6 &  52.7 &   2.5  & 17.4 & 46.7     &   8.0 &   -    &  -   \\
        VGGSound    & 39.4 &  88.6   &  39.4 & 1.8    &    50.0  &   3.4 &34.9 &  95.6  &  50.2 &  14.4 & 56.3 &  100.0   &   2.2 &  11.0 & 3.8   \\
        \bottomrule
        \end{tabular}
        \vspace{-0.5em}
    \caption{
      \textbf{Per-class mAP breakdown }
      For entries with `--' the test set does not contain any samples for that class.
    }
\label{tab:per-class}
\end{table*}

To better understand the strengths and weaknesses of our method, we report a performance breakdown by object class in \Cref{tab:per-class}.
We observe that the model obtains good results consistently for classes of large objects with a distinctive appearance (e.g. accordions and harps), while it is weaker for smaller objects such as oboes, or for objects that appear closely in numbers, like drums.

\paragraph{Comparison to audio-visual heatmaps.}

In~\Cref{tab:ciou} we compare the performance of our method trained on AudioSet to state-of-the-art sound source localisation methods. 
For a fair comparison to these methods, we convert the union of our predicted bounding boxes with confidence above a set threshold into a binary map, and use the latter as a pseudo-heatmap to use the same evaluation code.
Our approach outperform others for both class-agnostic single object localisation and for class-aware multi-object localisation, \textit{without} using audio signals during inference.

We note however that cIOU is not a very reliable metric for evaluating a detector (or even sound localizer) as it favours high recall over precision: by averaging this metric over all classes the most frequent ones (e.g. drums, guitars, pianos) dominate the metric. 
We therefore propose to the research community -- and report in this paper -- mean average precision (mAP) values as a more indicative metric.

\begin{table}[!tb]
\footnotesize
\centering
\setlength{\tabcolsep}{2pt}
    \begin{minipage}[b]{0.50\hsize}\centering

\begin{tabular}{l l c c}
\toprule
& & \multicolumn{2}{c}{\textbf{mAP$_{50}$}} \\
\cmidrule{3-4}
\# GT-cls. &  K  & VGGS & O.Images \\
\midrule
39   &   20   & 34.4  & 24.4  \\
39   &   30   & 35.1 & 25.1 \\
39   &   39   & 39.4 & \textbf{28.5} \\
39   &   50   & \textbf{41.0} & 27.5 \\
\bottomrule
\end{tabular}%
    \caption{
      \textbf{Number of clusters K.}
      Our method is relatively robust ($<\!5\%$ decrease in AP) to the number of self-labelling clusters.
      \label{tab:num_classes}
    }
    \end{minipage}
    \hfill
    \begin{minipage}[b]{0.46\hsize}\centering
        \begin{tabular}{l cc}
    \toprule
    & \multicolumn{2}{c}{\textbf{mAP$_{50}$}} \\
    \cmidrule{2-3}
    Matching   & VGGS & O.Images \\
    \midrule
     Hung.      & 39.4 &  28.5   \\
     Argmax         & 39.6 &  \textbf{30.1}   \\
     Manual         & \textbf{41.0} &  29.5   \\
    \midrule
     1-shot         & 36.4 & 25.1\\
     10-shot        & 37.1 & 25.8  \\
    \bottomrule
\end{tabular}%
    \caption{
      \textbf{Matching strategies.} 
      Even with 39 labels, our method performs accurately.
     \label{tab:cluster_matching}
     }
 \vspace{11pt}

    \end{minipage}
\vspace{-20pt}
\end{table}

\paragraph{Ablation: Number of clusters K.}

In~\Cref{tab:num_classes}, we perform an experiment varying the number of clusters $K$,
and as a consequence the number of object categories that the detector learns, while
keeping the test-set (containing 15 classes) fixed.
We observe that out method achieves reasonable performance for a wide range of $K$.
The performance is fairly stable when $K$ is more than the ground truth classes, and gradually decreases when fewer clusters are used.

\begin{figure*}[tbh!]
\centering
    \includegraphics[width=0.94\textwidth]{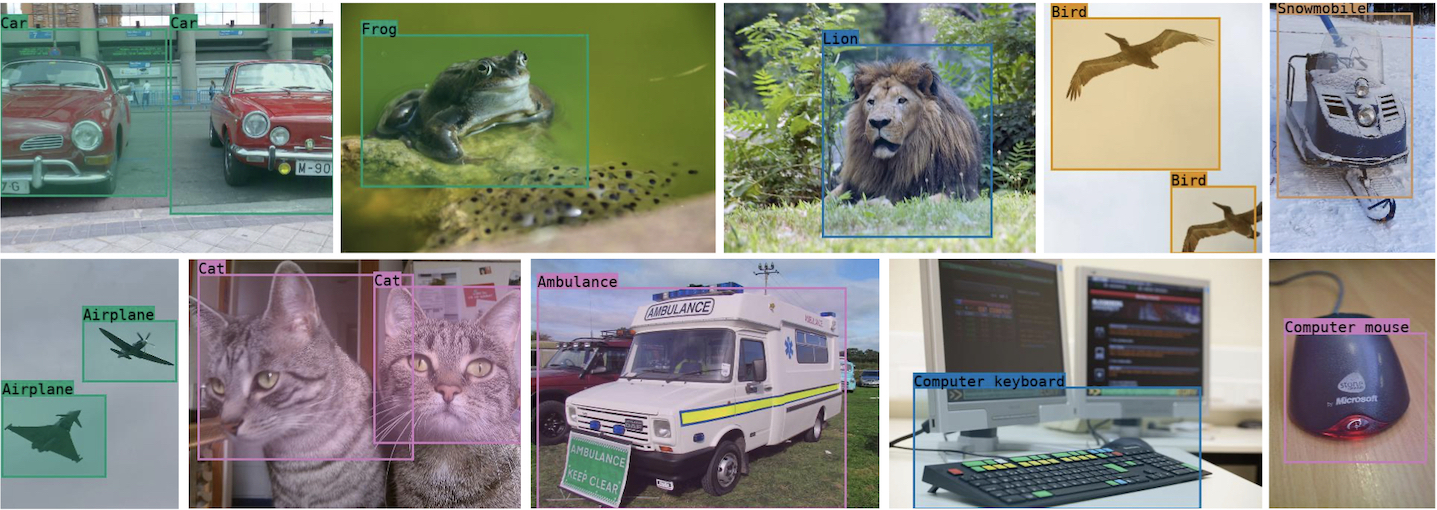}
            \vspace{-0.5em}
    \caption{\textbf{Object detection beyond musical instruments}. 
    Our proposed method can learn to accurately detect objects from more general categories, as long as they can be associated with a characteristic sound. The results shown here are from a model trained without labels directly on the full VGGSound dataset which includes 309 different video classes.
    Our method successfully learns to detect non-instrument objects, even in difficult multi-instance cases. 
     \label{fig:general_objects_main}}
          \vspace{-8pt}
\end{figure*}

\paragraph{Data-efficient detector alignment.}

In~\Cref{tab:cluster_matching} we conduct an investigation into the matching of the clusters to the ground truth labels. 
Recall that, for evaluation, we assign clusters to ground truth labels using the Hungarian algorithm. 
We can improve the computational efficiency of this step by using majority voting instead, which results in the same accuracy on VGGSound and a gain of 1.6\% on OpenImages.
Using a manual grouping strategy (see  \supp for details) yields another small boost.
We can also improve the statistical efficiency as follows.
While the evaluation protocol computes the optimal assignments assuming that all videos are labelled, in a real-world application we are interested in naming clusters with as little labelled data as possible.
To do this, we still use majority voting, but we assume that only the top $m$ videos per cluster (based on the strength of association) are labelled.
We find that even by just using $m=1$ (\ie, a \textit{total} number of 39 annotations), our method still achieves 37.1\% and 25.3\% on VGGSound and OpenImages. 
This is a 3\% drop compared to the Hungarian algorithm and can be further reduced to 2.3\% by using $m=10$ (\ie, 390 annotations).

\paragraph{Qualitative analysis.}
We show examples of successfully detected objects in challenging images in Fig.~\ref{fig:qualitative}, where we also include the outputs of the PCL baseline.
Although our model has not been manually shown any objects boundaries during training we see that it can learn very accurate boxes around them and that it can successfully identify multiple objects in complicated scenes. We provide further examples in \supp.

\paragraph{Towards general object detection.}
The results presented thus far have focused on subsets of common datasets with instruments solely to ensure comparability with prior works.
Since one main goal of self-supervised learning is to leverage the vast amount of unlabelled data, we wish to investigate how general and robust our proposed method when applied on a far larger scale.
For this, we increase our pretraining dataset by approximately $10\times$, simply by taking the whole of the VGGSound dataset, without any filtering. 
We set $K$ to $300$ and keep all training parameters the same; the result is an unsupervisedly trained object detector that can classify 300 pseudo-classes.
As before, we match these to the VGGSound labels with the Hungarian algorithm and select ten categories for which we have annotations in the OpenImages dataset (details in \supp).

\begin{table}[t]
\scriptsize
\centering
\begin{tabular}{l r r r}
    \toprule
    Class   &  AP$_{30}$ &  AP$_{50}$ & AP$_{[50:95:5]}$   \\
    \midrule
     Mean            &  45.6  & 24.4 &  6.5   \\
    \midrule
    Airplane         &  62.7  & 27.0 &  6,5   \\
    Ambulance        &  56.9  & 30.9 &  7.1   \\
    Bird             &  26.5  & 15.8 &  3.7   \\
    Car              &  29.8  & 18.4 &  5.1   \\
    Cat              &  67.7  & 28.0 &  7.7   \\
    Comp. Keyboard.  &  53.3  & 42.6 &  12.9   \\
    Comp. Mouse      &  35.9  & 25.4 &  8.8   \\
    Frog             &  43.5  & 19.5 &  4.7   \\
    Lion             &  34.1  & 22.2 &  4.9   \\
    Snowmobile       &  64.3  & 14.3 &  3.5   \\
    \bottomrule
\end{tabular}%
          \vspace{-5pt}
    \caption{
      \textbf{  Results on general object categories.}
     \label{tab:general_objects}
     }
              \vspace{-15pt}
\end{table}

In Fig.~\ref{fig:general_objects_main} we show qualitative results of some detections on OpenImages.
The numerical results are given in Table~\ref{tab:general_objects}. 
We find that even for objects that are deformable, such as cats, we get high AP$_{30}$ values of 67.7\% and that even objects that vary in shape, such as airplanes (see Fig.~\ref{fig:general_objects_main}, bottom-right), we achieve a good performances 62.7\%.
While the results for the AP$_{50:95:5}$ metric indicate that there is still room for improvement, these initial results show that leveraging larger and more diverse video datasets for self-supervisedly learning object detectors is a promising avenue. 
We note that, since minimal curation is performed on the training data, and we use a large number of different object categories in a noisy dataset, this training setting is very challenging. These results further highlight the potential of our proposed method.

\vspace{-5pt}
\section{Discussion}\label{s:conclusions}
\vspace{-5pt}
\paragraph{Limitations \& societal impact.}

We refer the reader to \supp for an extensive examination of failure cases. Detection failures manifest as
multiple instance grouping, missing instances or part-of-object detection, all well documented in the WSOD literature~\cite{ren20instance-aware}.
Another error is detecting wrong objects that often appear together with the objects of interest due to biases in the data (\eg~mouth regions as wind instruments). 
Wrong classification occurs often because i) visually similar classes are confused - \eg~Horn and Trumpet; ii) of incorrect semantic matching; iii) confusion due to the object’s orientation –
\eg~a vertical violin confused for a cello. 
Regarding training, there are likely further optimizations possible, such as a more end-to-end design.
As for most unsupervised methods, a downside of our approach compared to supervised detection is the reduced human control on the learned concepts, which may warrant additional manual validation before deployment.

\vspace{-5pt}
\paragraph{Conclusion.}
We have presented a method for training strong object detectors purely with self-supervision by watching unlabelled videos.
We demonstrated that our best models perform better than a heatmap-based methods while not requiring and audio and better than weakly supervised baselines, even after curating the dataset to filter out noisy samples for training the latter. 
We have also addressed one shortcoming of using the Hungarian algorithm for evaluation by showing that data-efficient alignment of self-supervised detectors is possible with as little as one image per pseudo-label.
Finally, we applied our method to domains beyond musical instruments and found that it can learn reasonable detectors in this much less curated setting, paving the way to general self-supervised object detection.

\vspace{-5pt}
\subsection*{Acknowledgements}
\vspace{-5pt}
\noindent Y.M.A is thankful for MLRA funding from AWS. 

{\small\bibliographystyle{ieee_fullname}\bibliography{shortstrings,vgg_local,vgg_other,refs,vedaldi_specific,vedaldi_general}}

 \clearpage
 \newpage
 \renewcommand{\thefigure}{A.\arabic{figure}} %
\setcounter{figure}{0} 
\renewcommand{\thetable}{A.\arabic{table}}
\setcounter{table}{0} 

\bigskip
{\noindent \huge \bf {Appendix}} \\

\appendix

This Appendix provides: additional qualitative results including failure analysis (Sec.~\ref{app:sec:qualitative}),
further implementation details (Sec.~\ref{app:sec:implementation})
and label mapping details (Sec.~\ref{app:sec:dataset}).

\section{Qualitative analysis \& failure cases}
\label{app:sec:qualitative}

\begin{figure*}[tb]
\centering
    \includegraphics[width=0.95\textwidth]{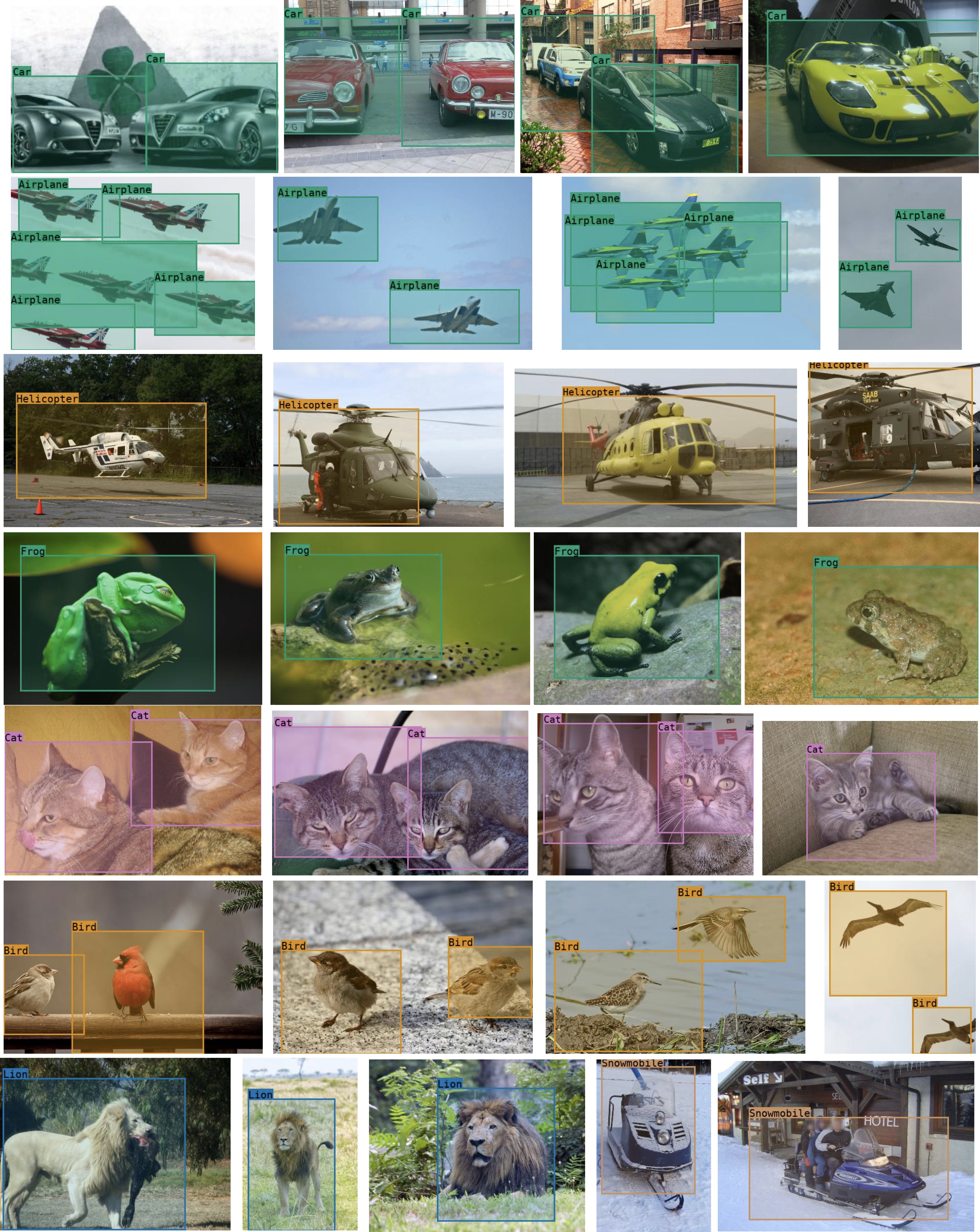}
    \caption{\textbf{Object detection beyond musical instruments}. 
    Our proposed method can learn to accurately detect objects from more general categories, as long as they can be associated with a characteristic sound. The results shown here are from a model trained without labels directly on the full VGGSound dataset which includes 309 different video classes.
     \label{fig:beyond_instruments}}
\end{figure*}

\begin{figure*}[tb]
\centering
    \includegraphics[width=0.97\textwidth]{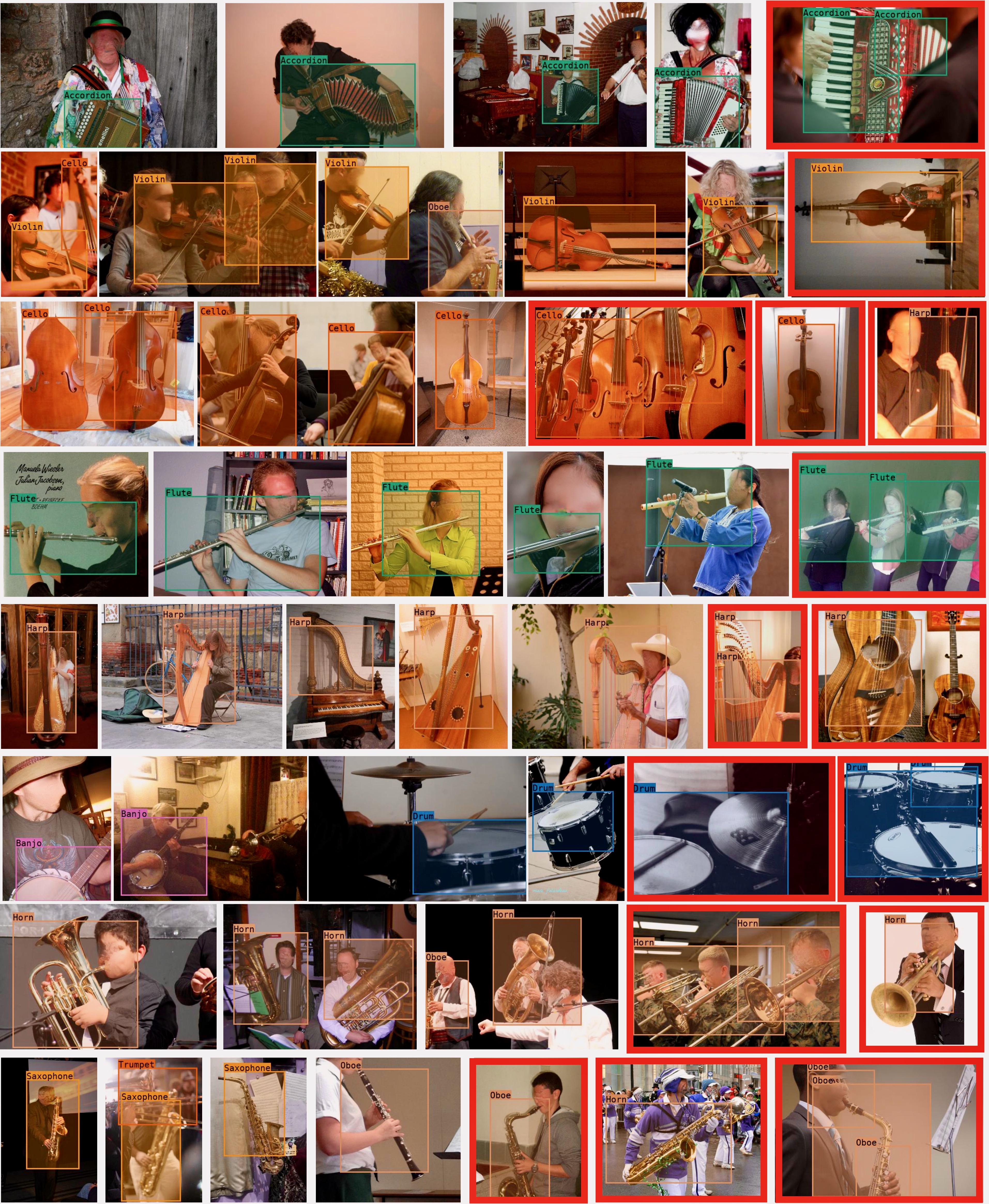}
    \caption{\textbf{Additional qualitative examples and failure cases.} The images highlighted in red contain failure cases.  
     \label{fig:qual_failure}}
\end{figure*}

In Figure \ref{fig:beyond_instruments} we show more qualitative results for the general object detection experiment.
In Figure~\ref{fig:qual_failure} we show some more examples of our trained detector's output on detecting musical instruments on the OpenImages test set, including failure cases. 
As noted in the main paper, the model's detections are very precise, especially for visually distinct categories such as Accordion, Harp, Cello etc.

The failure cases can be attributed to either incorrect box predictions or misclassifications. 
Box-prediction failure modes most commonly manifest as i) grouping multiple instances into the same detection, ii) missing instances or iii) detection of only parts of an object. All those issues are common and well documented in the WSOD literature~\cite{ren2020instance}.
We observe that in the case of objects that appear with high granularity -- e.g. drums -- although qualitatively the detections seem reasonable, the multiple instance grouping problem is prominent; this explains our model's low performance by quantitative measures (mAP) for these classes. 

Another error is detecting parts of wrong objects that often appear together with the objects of interest due to biases in the data. For example sometimes mouth regions are detected as wind instruments. This is expected when using the audio-visual correspondence as the training signal. 

Wrong classification on the other hand occurs often because i) visually similar classes are confused - e.g. Saxophone and Oboe, Horn and Trumpet; ii) of incorrect cluster-to-ground-truth-label matching, e.g. cluster with high Tombone purity matched to Horn; iii) sometimes the model confuses the class due to the object's orientation -- for example a vertical violin is detected as a cello or vice versa a horizontal cello is detected as violin.

\section{Additional implementation details}
\label{app:sec:implementation}
\paragraph{Visual \& audio encoder architecture.}
For the visual and audio encoders ($q^v,q^a$) we use the AVOL-Net~\cite{arandjelovic18objects} architecture. 
Each network is a 8-layer CNN with an architecture similar to VGG-M. For a typical  $224 \times 224 \times 3$ RGB image input, the video encoder outputs a $14 \times 14$ visual feature map.
The inputs to the audio encoder are $257 \times 200$ dimensional log-mel spectrograms, extracted from 1 second of audio sampled at 24Hz. The output of the encoder is average pooled, resulting in a single audio feature vector. 

\paragraph{Localisation and classification heads.}
As localisation $f^{\{v,a\}}$ and classifier $g^{\{v,a\}}$ networks we use for each modality a 2-layer MLP with a hidden dimension of $512$. 
The MLPs do not share weights and are applied on top of the common representations extracted by the video and audio encoders.
The visual feature maps are average-pooled before being input to the classification network $g^{v}$, while $f^{v}$ is applied directly on the spatial feature map, as discussed in Section 3.1 of the main paper.  
The output dimension for the localisation networks (i.e. the common embedding dimension) is $128$, while the audio and video classifier heads each output $K$ values, one for every cluster. 

\paragraph{Detector backbone.} As the backbone for the Faster R-CNN detector we use a ResNet50(1x)~\cite{He15,chen2020simple} model pre-trained without labels using the SimCLR~\cite{chen2020simple} method on ImageNet. We use the pretrained weights\footnote{\url{https://github.com/google-research/simclr}} released by the authors of \cite{chen2020simple}.
We apply a Feature Pyramid Network (FPN)~\cite{lin2017feature} on top of the activations of different backbone layers extracting visual features at five spatial scales, \ie /4, /8, /16, /32, and /64 of the input resolution.
Feature descriptors are extracted for every proposal with ROI-align pooling~\cite{he17mask}. 

\paragraph{Detector RPN training.}
The box proposals $m \in M(v)$ used by
the Faster R-CNN detector are produced by a Region Proposal Network (RPN).
The RPN extracts a set of anchors $n \in N(v) \subset \Omega^2$ at three aspect ratios (0.5, 1.0 and 1.5) on each FPN output scale.
The extracted number of anchors depends only on the input image dimensions.

The RPN contains 2 MLP heads in order to model
$
o(n) = f_\text{rpn}^\text{obj}(n|v)
\in \{0,1\}
$, a binary objectness label indicating whether anchor $n$ contains an object or background, 
and
$
m(n) = f_\text{rpn}^\text{loc}(n|v)
\in \mathbb{R}^4
$, the bounding box proposal resulting from anchor $n$.

The RPN heads are trained jointly with the detector heads, using the input box annotations and ignoring the class labels:
if $n^* = \argmax_{n\in N(v)} \operatorname{IoU}(n,t^*)$ is the anchor that matches the pseudo-ground truth bounding box $t^*$ the best, one optimizes:
\begin{multline*}
\mathcal{L}_\text{rpn}(v,t^*)
=
\mathcal{L}_\text{reg}(m(n^*),t^*)
+
\mathcal{L}_\text{obj}(o(n^*),1) \\
+
\sum_{a \in A(v): \operatorname{IoU}(n,t^*) < \tau} 
\mathcal{L}_\text{obj}(o(n),0).
\end{multline*}
Here $\mathcal L_\text{reg}$ is the $L^1$ loss for the bounding box corner coordinates and $\mathcal{L}_\text{obj}$ a binary cross-entropy loss.
Similarly to the detector loss, the RPN loss requires the best anchor
$n^*$ to match bounding box $t^*$ and to have a high objectness score, while reducing the objectness scores of anchors $n$ that are a bad match.

\paragraph{Baselines.}

There is currently no prior work on learning an object detector for multiple object classes without any supervision.
Instead, we compare against weakly-supervised detectors (hence using image-level labels) and unsupervised localisation methods that only produce heatmaps (not detections).

\textbf{Weakly-supervised detection.}

For weakly-supervised detection, we consider PCL~\cite{tang20pcl:}, the strongest such baseline for which we could find an implementation.

Image-level labels are obtained from the corresponding dataset: for AudioSet the 13 labels of the training set are used,
and for VGGSound we manually merge the 39 sub-classes to equivalent 15 classes (e.g.,~electric guitar, acoustic guitar to ``Guitar'' etc.); see \supp for full details.
However, we have found that training PCL directly on the same data as our method (i.e., random clips from the VGGSound and AudioSet-Instrument subsets) does not work, likely due to the high amount of noise present in the labels (e.g., several videos are be labelled with an instrument, which is however not visible at all).
To avoid this issue, we further preprocess the training data with a supervised instrument detector~\cite{gao19co-separating} and only retain frames where at least one relevant detection is found.
This of course gives an ``unfair'' advantage to the baseline, but it is necessary to be able to use it at all.

\textbf{Heatmap-based localisation.}

For our second baseline, we consider localisation methods that, similarly to us, use cross-modal self-supervised learning.
The state-of-the art DSOL method of~\cite{hu2020discriminative} is the most relevant, as it produces a heatmap roughly localizing the objects and produces class pseudolabels.
While DSOL does not use image-level labels like PCL, it does use audio during inference, and thus strictly more information than our method (which performs localisation only in the visual domain).

\textbf{Region proposals.}
Finally, we also compare against other baselines such as simply predicting a large centered box and class-agnostic region proposal methods such as selective search, and using a RPN obtained from supervised training on COCO~\cite{lin14microsoft}.

\paragraph{Training details.}
For all the experiments we pre-train the video and audio backbones using the localisation objective only and average-pooling instead of max-pooling on the heatmap. The pre-training is performed on the Audioset-Instruments training set for 230 epochs on 64 GPUs with a batch size of 12 per GPU, using the Adam optimizer with a constant learning rate of $6.4 \times 10^{-4}$ and standard hyperparameters. 
The learning rate is initially set to $1 \times 10^{-5}$ and warmed up gradually until reaching the constant value after 10 epochs.
The joint localisation and clustering training is performed on 16 GPUs with a batch size of 16 per GPU and SGD optimization with a constant learning rate of $0.005$ and momentum $0.9$, for a further 300 epochs.
We set hyperparameter $\lambda$ to 0.5. 
The temperature for the contrastive learning is learned as a single scalar weight. 
The detector training is performed on 16 GPUs with a batch size of 12, SGD optimization and learning rate $0.008$ for 100 epochs. 
The total training takes approximately 3 days in this setting.

\textbf{Training resolutions.}
We train our method on random square crops of 224 pixels after resizing to 256 pixels. 
During the training of the detector, we take random 224 crops and obtain the the self-supervised bounding boxes on-the-fly from our pretrained model, which are scaled and used to train the detector at the larger detector resolution. %

\textbf{Detector warm-up.}
We warm-up the Faster R-CNN detector by training in a class-agnostic manner for 20 epochs.
This gives the RPN (which is randomly initialized) an opportunity to learn sufficiently stable bounding box proposals; we then switch to full class-aware supervision.
We found that this leads to more robust convergence compared to training with pseudo-labels from the start.

\textbf{Backbone pretraining.}
We also found it beneficial to pretrain the localizer backbone using only the localisation loss on the full AudioSet-Instruments dataset, and the detector backbone using self-supervised SimCLR~\cite{chen2020simple} on  ImageNet (note that the DSOL baseline uses instead \emph{supervised} ImageNet pretraining for the backbone).

\section{Additional label mapping details}
\begin{table}[!t]
    \centering
    \setlength{\tabcolsep}{10pt}
    \begin{tabular}{ll}
    \toprule
    \textbf{VGGSound label}  & \textbf{OpenImages label} \\
    \textbf{ (representative) } & \textbf{ (test)} \\
    \midrule
    airplane flyby       &    Airplane    \\
    ambulance siren       &   Ambulance   \\
    mynah bird singing   &    Bird    \\
    racing car           &    Car    \\
    cat purring          &    Cat    \\
    typing on computer keyboard   &    Computer Keyboard   \\
    mouse clicking      &    Computer Mouse    \\
    frog croaking        &    Frog    \\
    helicopter           &    Helicopter    \\
    lions roaring        &    Lion    \\
    driving snowmobile   &    Snowmobile    \\
    \bottomrule
    \end{tabular}%
    \caption{
      Mapping of VGGsound to OpenImages labels for qualitative evaluation on non-instrument object categories shown in Figure~\ref{fig:beyond_instruments}.
      \label{tab:supp:vggsound_classes_300}
    }
\end{table}

\label{app:sec:dataset}
In Table~\ref{tab:supp:vggsound_classes} we show the full list of VGGSound video categories used for training our models.
We use only samples from the 39 \textit{mappable} classes for the detector training stage if not stated otherwise.

The 15 \textit{test} labels shown are the musical instrument labels found in the OpenImages dataset. As explained in the main paper, the Audioset and VGGSound test sets that we use for evaluation have been pseudo-labelled with high confidence detections from a supervised detector trained on OpenImages by \cite{gao19co-separating}, therefore these 15 test labels are common among all our test sets.

We use the same colour to show the manual grouping of VGGSound labels that we use for some of the baseline and ablation experiments. The many-to-one grouping of the \textit{mappable} into \textit{representative} labels is used for training the PCL WSOD baseline and also for evaluating the ``manual'' matching strategy in Table 6 of the main paper.
We note that our proposed method \textbf{does not} use this mapping, but instead uses the Hungarian algorithm for matching our learned clusters directly to the \textit{representative} classes.
On the same table we also show the constant mapping of \textit{representative} VGGSound labels to the OpenImages test labels that is necessary for evaluation. The majority of the instrument categories are matched exactly, except for the `Guitar' and `Drum' test labels for which we had to choose one of the candidates from the VGGSound guitar and drum subsets; we chose ``playing electric guitar'' and ``playing snare drum'' as the representative labels respectively. Note that since the Hungarian matching only matches one cluster to every test label, only choosing one \textit{representative} class, harms our detectors, which are in fact more specialised. For example instances of acoustic guitar that may be correctly detected and not labelled as electric guitar will not be taken into account, in effect lowering our method's measured performance.

Similarly in Table~\ref{tab:supp:vggsound_classes_300} we show the \textit{representative} to \textit{test} mapping used for evaluation and visualisation in the non-instrument object detection experiment.

\begin{table}[]
\footnotesize
\centering
        \begin{tabular}[h]{cccc}
    \toprule
    & & \multicolumn{2}{c}{\textbf{mAP$_{50}$}} \\
    \cmidrule{3-4}
    $\beta$ & \#boxes & VGGS  & O.Images \\
    \midrule
    0.7      &  single  &  39.4  & 28.5 \\
    0.8      &  single  &  36.6  & 29.4 \\
    0.9      &  single  &  35.8  & 28.8 \\
    0.7-0.9  &  single  &  37.5  & \textbf{29.4} \\
    0.7-0.9    &  multi  &  \textbf{38.0}  & 29.3 \\
    \bottomrule
    \end{tabular}%
    \caption{
      \textbf{Ablation of hyperparameter {\boldmath$\beta$}}, which controls the relative width of the bounding box; \textit{multi} denotes the use of multiple self-labelled boxes per sample. 
     Overall the method is fairly stable with respect to the choice of this parameter, however sampling $\beta$ from a range (0.7-0.9) obtains a better balanced performance. Moreover we do not observe any substantial improvement from using multiple boxes.
    }
\label{tab:beta_ablation}

\end{table}

\begin{table*}[t]
  \centering
   \setlength{\tabcolsep}{20pt}
\resizebox{0.8\textwidth}{!}{%
\begin{tabular}{llll}
    \toprule

   \textbf{VGGS instrument classes (50)} & \textbf{mappable (39)} & \textbf{representative (15)}  & \textbf{test (15)} \\
    \midrule
  
  playing accordion      &                  {\color{red}     playing accordion }     &                    {\color{red} playing accordion }      &   {\color{red} Accordion}    \\
playing double bass      &                  {\color{Orange}   playing double bass }     &                   {\color{Orange}  playing cello }     &       {\color{Orange}  Cello}       \\
playing bassoon      &                      {\color{OliveGreen}   playing bassoon }     &                       {\color{Mulberry}  playing flute }     &      {\color{Mulberry}  Flute}        \\
playing cello      &                        {\color{Orange}   playing cello }     &                        {\color{RawSienna}   playing electric guitar}      & {\color{RawSienna}  Guitar }      \\
playing clarinet      &                     {\color{OliveGreen}   playing clarinet }     &                      {\color{Violet}  playing harp  }    &   {\color{Violet} Harp }       \\
playing cornet      &                       {\color{GreenYellow}   playing cornet  }    &                       {\color{OliveGreen}   playing oboe}      &  {\color{OliveGreen}  Oboe }      \\
playing flute      &                        {\color{Mulberry}   playing flute   }   &                        {\color{NavyBlue}   playing piano  }    &   {\color{NavyBlue} Piano }      \\
playing acoustic guitar      &              {\color{RawSienna}   playing acoustic guitar }     &              {\color{Tan}   playing saxophone}      &  {\color{Tan}  Saxophone }      \\
playing bass guitar      &                   {\color{RawSienna}  playing bass guitar  }    &                   {\color{Lavender}  playing trombone }     &  {\color{Lavender} Trombone}       \\
playing electric guitar      &               {\color{RawSienna}  playing electric guitar }     &                {\color{GreenYellow} playing trumpet  }    &   { \color{GreenYellow}Trumpet}       \\
tapping guitar      &                       {\color{RawSienna}   tapping guitar }     &                        {\color{Magenta}  playing violin  }    &  {\color{Magenta}  Violin }      \\
playing harp      &                         {\color{Violet}   playing harp  }    &                         {\color{SeaGreen}   playing banjo }     &  {\color{SeaGreen}  Banjo }       \\
playing harpsichord      &                  {\color{NavyBlue}   playing harpsichord  }    &                  {\color{LimeGreen}   playing harmonica }     &   {\color{LimeGreen} Harmonica}       \\
playing oboe      &                         {\color{OliveGreen}   playing oboe  }    &                          {\color{Goldenrod}  playing french horn }     &   {\color{Goldenrod} Horn }      \\
playing electronic organ      &             {\color{NavyBlue}   playing electronic organ  }    &             {\color{Sepia}   playing snare drum }     &  {\color{Sepia}  Drum  }     \\
playing hammond organ      &               {\color{NavyBlue}    playing hammond organ }     &           \\
playing piano      &                        {\color{NavyBlue}   playing piano  }    &           \\
playing saxophone      &                    {\color{Tan}   playing saxophone  }    &           \\
playing trombone      &                    {\color{Lavender}    playing trombone  }    &           \\
playing trumpet      &                      {\color{GreenYellow}   playing trumpet  }    &           \\
playing ukulele      &                     {\color{RawSienna}    playing ukulele  }    &           \\
playing violin      &                       {\color{Magenta}   playing violin   }   &           \\
playing mandolin      &                     {\color{SeaGreen}   playing mandolin }     &           \\
playing banjo      &                        {\color{SeaGreen}   playing banjo  }    &           \\
playing harmonica      &                    {\color{LimeGreen}   playing harmonica  }    &           \\
playing french horn      &                  {\color{Goldenrod}   playing french horn  }    &           \\
playing drum kit      &                    {\color{Sepia}    playing drum kit  }    &           \\
playing bass drum      &                    {\color{Sepia}   playing bass drum }     &           \\
playing snare drum      &                   {\color{Sepia}   playing snare drum  }    &           \\
playing tabla      &                       {\color{Sepia}    playing tabla  }    &           \\
playing bongo      &                       {\color{Sepia}    playing bongo }     &           \\
playing tambourine      &                  {\color{Sepia}    playing tambourine }     &           \\
playing timpani      &                     {\color{Sepia}    playing timpani  }    &           \\
playing tympani      &                     {\color{Sepia}    playing tympani }     &           \\
playing timbales      &                     {\color{Sepia}   playing timbales  }    &           \\
playing congas      &                       {\color{Sepia}   playing congas   }   &           \\
playing djembe      &                       {\color{Sepia}   playing djembe  }    &           \\
playing cymbal      &                      {\color{Sepia}    playing cymbal  }    &           \\
playing gong      &                        {\color{Sepia}    playing gong }     &           \\
playing didgeridoo      &           \\
playing bugle      &           \\
playing shofar      &           \\
playing glockenspiel      &           \\
playing vibraphone      &           \\
playing marimba      &           \\
playing bagpipes      &           \\
playing theremin      &           \\
playing steel guitar      &           \\
playing sitar      &           \\
playing erhu      &           \\

    \bottomrule
\end{tabular}%
}
    \caption{
      \textbf{VGGSound instrument catergories and labels mapping}.
      The 15 \textit{test} labels are the OpenImages instrument labels that all our tests sets are labelled with.
      The 39 \textit{mappable} labels comprise the VGGSound subset that can be roughly mapped to those 15 test labels -- this mapping is necessary for evaluation.
      The \textit{representative} classes are the ones that we use to link VGGSound labels with the 15 test labels for evaluation. 
    Labels manually matched by us are shown with the same colour.
    }\label{tab:supp:vggsound_classes}
\end{table*}

\section{Additional experiments}

\paragraph{Ablation: Thresholding parameter.}
In~\Cref{tab:beta_ablation} we investigate the influence of the hyper-parameter $\beta$ and the number of extracted target boxes we use for training the detector.
From Eq.~\eqref{e:heatmap_threshold}, a smaller $\beta$ makes the heatmaps more focused and specific.
With regards to this parameter, we find somewhat inverse trends for VGGSound vs OpenImages, where smaller $\beta$ yields better results for the former and larger $\beta$ for the latter.
We find that a good balance in terms of performance can be achieved by sampling $\beta$ randomly from a range, as  over-specific boxes for some images and under-specific ones are successfully combined during detector training.
Even when extracting multiple boxes for training the detector, we find our method performs similarly well to when only extracting a single box.

 \clearpage
 \newpage

\end{document}